%% file: fruct_main.tex
\documentclass[conference]{IEEEtran}
\renewcommand{\figurename}{Fig.}

\usepackage[dvips]{graphicx}
\usepackage{cite}
\usepackage{amsmath,amssymb,amsfonts}

\usepackage{tabu}
\usepackage[linesnumbered,ruled,vlined]{algorithm2e}
\usepackage{algpseudocode}
\usepackage{comment}
\usepackage{balance}
\usepackage[labelsep=period,font=footnotesize]{caption}
\SetKwInput{KwInput}{Input}                
\SetKwInput{KwOutput}{Output}              
\usepackage{parskip}
\makeatletter

\usepackage{subcaption}

\usepackage{textcomp}
\usepackage{xcolor}
\usepackage{url}
\usepackage{multirow}

\begin{document}
\title{Depth and Image Fusion for Road Obstacle Detection Using Stereo Camera}
\date{}

\author{
\IEEEauthorblockN{Oleg Perezyabov\textsuperscript{1}, Mikhail Gavrilenkov\textsuperscript{2} and Ilya Afanasyev\textsuperscript{3}}
\IEEEauthorblockA{
\textsuperscript{1,2}\textit{Huawei St. Petersburg Research Center}  \\
\textsuperscript{3}\textit{Independent Researcher} \\
\textsuperscript{1}perezyabov.oleg1@huawei.com, 
\textsuperscript{2}gavrilenkov.mikhail1@huawei.com,
\textsuperscript{3}ilya.afanasyev@gmail.com
}}

\maketitle

\input{chapters/abstract}

\input{chapters/introduction}

\input{chapters/related_work}

\input{chapters/methodology}

\input{chapters/setup}

\input{chapters/conclusion}

\bibliography{fruct_main}
\bibliographystyle{IEEEtran}

\end{document}

%% file: chapters/abstract.tex
\begin{abstract}

This paper is devoted to the detection of objects on a road, performed with a combination of two methods based on both the use of depth information and video analysis of data from a stereo camera. Since neither the time of the appearance of an object on the road, nor its size and shape is known in advance, ML/DL-based approaches are not applicable. The task becomes more complicated due to variations in artificial illumination, inhomogeneous road surface texture, and unknown character and features of the object. To solve this problem we developed the depth and image fusion method that complements a search of small contrast objects by RGB-based method, and obstacle detection by stereo image-based approach with SLIC superpixel segmentation. We conducted experiments with static and low speed obstacles in an underground parking lot and demonstrated the successful work of the developed technique for detecting and even tracking small objects, which can be parking infrastructure objects, things left on the road, wheels, dropped boxes, etc.

\end{abstract}

%% file: chapters/introduction.tex
\section{Introduction}\label{sec1}

The intelligent transport monitoring system for roads, parking lots and highways must detect the sudden appearance of obstacles on the road that can lead to an emergency. Frequently objects, boxes, loads, etc. that have fallen from passing vehicles can become a source of such obstacles (that are especially dangerous on busy highways), as well as objects moving at low speed (such as animals).

According to European Agency for Safety and Health at Work (EU-OSHA), around a third of the deaths of people in workplace accidents in the EU are related to transport \cite{copsey2010review}. What is more, within the accidents that involve the people the reasons can be objects falling from vehicles. The EU-OSHA review of accidents and injuries to road transport drivers notes that "If a cargo is not adequately secured, it can be a danger to the driver and to others: the cargo can fall off the vehicle and form an obstacle that in turn may hurt or kill the driver or other road users... During strong braking or a crash the risk of cargo falling off the vehicle is increased" \cite{copsey2010review}.

For this purpose, monitoring systems used at intersections or parking lots, consisting of surveillance cameras and, sometimes, depth sensors, it is logical to supply with the specific functions for analyzing the traffic situation, timely detection of both static and dynamic obstacles in real time, and warning drivers. It is assumed that sensor data processing can take place both on sensor systems if they have sufficient computing resources (like edge computing), and in the cloud if there is a high-speed data transmission channel (for cloud computing), and, possibly, on the monitoring system server (if a computing cluster is provided).

It should be noted that different monitoring systems may have different restrictions on obstacle detection. So, Doppler millimeter radars cannot register static objects, lidar systems create dense point clouds that are difficult to process in real time, and also have a limited working distance (about 200 meters). Artificial intelligence systems with the detection and classification of objects can be inefficient, since neither the time of the appearance of the object, nor its nature, size, color, shape, etc. are known in advance. Moreover, it is difficult to even prepare the necessary data set for training a neural network, plus the performance of a ML/DL-based system for real-time high-resolution image processing may be questionable.

In this work, a stereo camera system was chosen for the study, since it meets such requirements as, on the one hand, it provides depth information by computing the disparity map (and, consequently, the distance to the object), on the other hand, the use of a video stream from one of stereo cameras gives us the opportunity to apply computer vision algorithms for object detection and tracking. Merging sensory information for both depth and vision yields more accurate object detection, positioning and characterization. Therefore, this article focuses on a depth and image fusion algorithm that takes advantage of the dual detection of road obstacles through the different processing channels of stereo and RGB information.

It should be noted that for monocular obstacle detection on the road with frame-by-frame image analysis, we used the OpenCV graph segmentation methodology with the following pre-processing: converting an RGB image to HSV with subsequent saturation increase, median filtering, erosion and dilation to remove noise. For depth-based obstacle detection, we use the recovery of disparity map from stereo (performed automatically by selected stereo camera software Stereolabs Zed 2 and Intel RealSense SDK), getting a dense point cloud and then applying the superpixel algorithm, to speed up point cloud processing and noise reduction to reduce the influence of background noise, which fluctuates greatly with time and illumination.

As restrictions on the experiments, we would note that the area-of-interest for the underground parking lot was investigated with several types of objects such as boxes of different sizes, the orange-white traffic cones, and a person in the camera's field of view). Since we used Stereolabs Zed 2 and Intel RealSense stereo cameras available on the market and could not change the baseline (the distance between the cameras), the obstacle detection distance is limited to 10-15 meters. Due to significant stereo camera noise in the raw data streams from the SDK, we could not distinguish objects smaller than 10x10x10 cm without special RGB camera processing.

As the main research contribution of this article, we would like to highlight the following:
\begin{enumerate}
    \item A combined algorithm for detecting obstacles on the road using dual channels for processing stereo and mono information, followed by depth and image fusion with specially selected metrics.
    \item An improved technique for detecting obstacles on the road applying a monocular camera with frame-by-frame image analysis, associated with graph segmentation and special pre-processing, using saturation increase, median filtering, erosion and dilation, rather than using neural networks and a priori known scene (i.e. a pre-recorded scene with an obstacle-free road).
\end{enumerate}

The rest of the paper is organized as follows: The Section \ref{sec:related_work} considers research papers on road obstacles detection. The detailed methodology and implementation of our proposed algorithm is explained with examples in the Section \ref{sec:methodology}. The Section \ref{sec:setup} describes the experimental setup, tests and results. Finally, we discuss and conclude in the Section \ref{sec:conclusion}.

%% file: chapters/related_work.tex
\section{Related Work}
\label{sec:related_work}

The problem of obstacle detection on the road is devoted to many different studies and reviews \cite{yu2020study, rateke2020passive, hu2020survey}. Part of the research is devoted to the real-time detection of obstacles from on-board sensors of intelligent transport systems, driver assistance systems and automated vehicles \cite{godha2017road,yu2020study,badue2021self}. 
And as noted in \cite{yu2020study}, the most common autonomous support functions for commercially available premium piloted vehicles are autonomous highway driving, semi-autonomous parking, and braking.

The methods that apply image processing are various. 
For example, the method described in article \cite{godha2017road} uses morphological filtering of camera frames to detect obstacles in front of a moving vehicle in order to warn the driver about avoiding collisions (in this case, other road users can often be considered as an obstacle).
In our case, we use traffic monitoring with a stationary sensor (i.e. stereo camera).

Frequently, the detection of known classes of objects on the road for transport tasks is carried out using ML/DL-based approaches \cite{dhiman2019pothole,rateke2020road}. But, as we have already noted, if the type, shape and moment of an obstacle's appearance on the road is not known in advance, the efficiency of applying neural networks can be questionable. In addition, assembling a dataset for neural network training can be a difficult task.

Often in studies, to detect obstacles cameras and computer vision algorithms are used \cite{dhiman2019pothole,rateke2020passive}. However, sometimes depth sensors such as stereo cameras \cite{rateke2020road,badue2021self}, LiDARs \cite{chen2017lidar,yu2020study} and millimeter wave radars \cite{zhang2019radar,yu2020study} can be applied as well. 

In our study, we use a stereo camera, as it combines the advantages of using depth information (and therefore the distance to the object) and the ability to process the video stream using computer vision algorithms.

%% file: chapters/methodology.tex
\section{Methodology and Algorithm Implementation}
\label{sec:methodology}

Obstacles detection experiments have been performed on dataset collected by our own in underground parking scenario (see the Section \ref{sec:setup}). For ZED2 camera (Fig. \ref{fig:fig13}a) the dataset includes video sequence from left and right RGB cameras and disparity map, retrieved via stereo camera SDK. For Intel RealSense Camera (Fig. \ref{fig:fig13}b) the dataset consisted of left and right grayscale images, RGB image for color data and disparity map, obtained by camera SDK.
The dataset scenario is as follows: parking spaces and parking  passages with the boxes of 10x10x10 cm and 20x20x20 cm being thrown or brought by human in the camera field of view.

\begin{figure*}[!htbp]
\captionsetup{font=scriptsize}
\centering
\includegraphics[width=\textwidth]{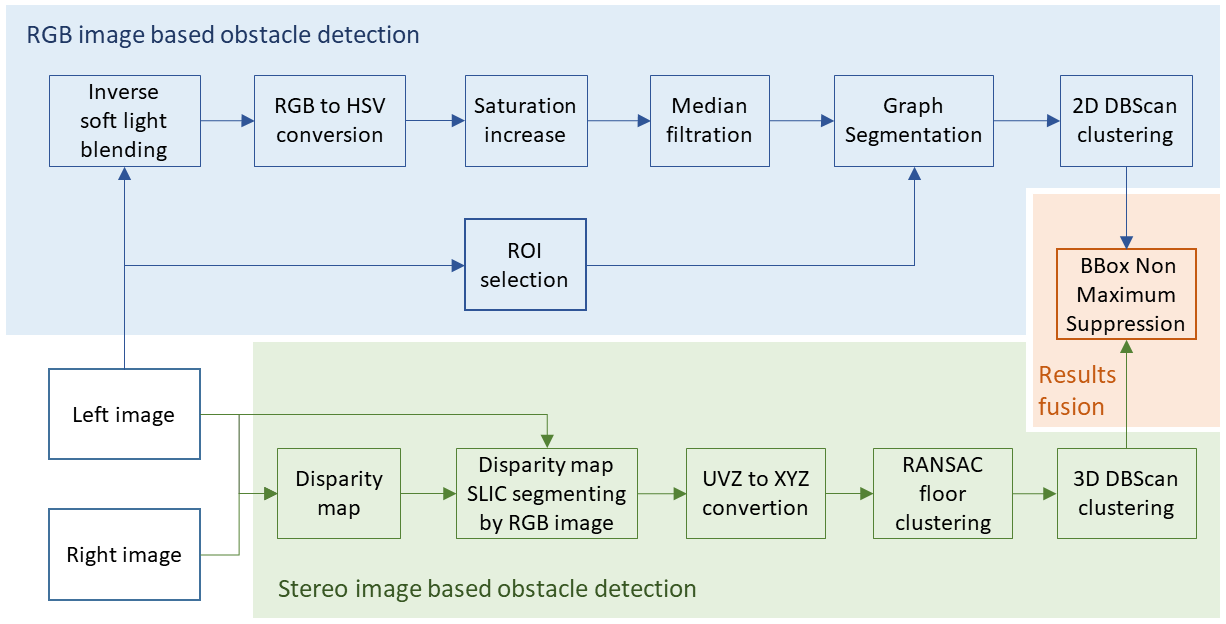}
\caption{\fontsize{8}{12}\fontfamily{ptm}\selectfont The methodology of depth  and image fusion for road obstacle detection using stereo camera}
\label{fig:method}
\end{figure*}

The proposed method of depth and RGB image fusion for road obstacle detection, and its detailed workflow is demonstrated in Figure \ref{fig:method}.

\subsection{RGB image-based obstacle detection}

We are faced with the task of searching for all the obstacles that can lead to traffic accidents. Therefore, our goal is to find both objects that were already on the ground when the camera was turned on, and objects that appeared in the field of view later. That is why we cannot apply a method that uses the so-called key frame, that is, a reference image in which we are sure that there are no obstacles. Also, we cannot implement a method that requires a priori information about the foreground and background, because we do not have enough information about the possible obstacle and its properties. Among the widespread image segmentation methods, the graph segmentation described in \cite{felzenszwalb2004efficient} seems to be very promising for solving our task, since it segments the image both in terms of spatial and color Euclidean distance (equation \ref{eq1}).

\begin{equation}\label{eq1} 
\begin{gathered}
dist(p_{i},p_{j}) = \sqrt{(x_{i}-x_{j})^{2}+(y_{i}-y_{j})^{2}+...} \\
\sqrt{...+(r_{i}-r_{j})^{2}+(g_{i}-g_{j})^{2}+(b_{i}-b_{j})^{2}}
\end{gathered}
\end{equation}

where $x$ and $y$ are 2D coordinates of $p_{i}$ and $p_{j}$ image points, and $r$, $g$, $b$ are the intensities of the red, green, and blue pixels, respectively.

\begin{figure}[!htbp]
    \centering
    \includegraphics[width=\linewidth]{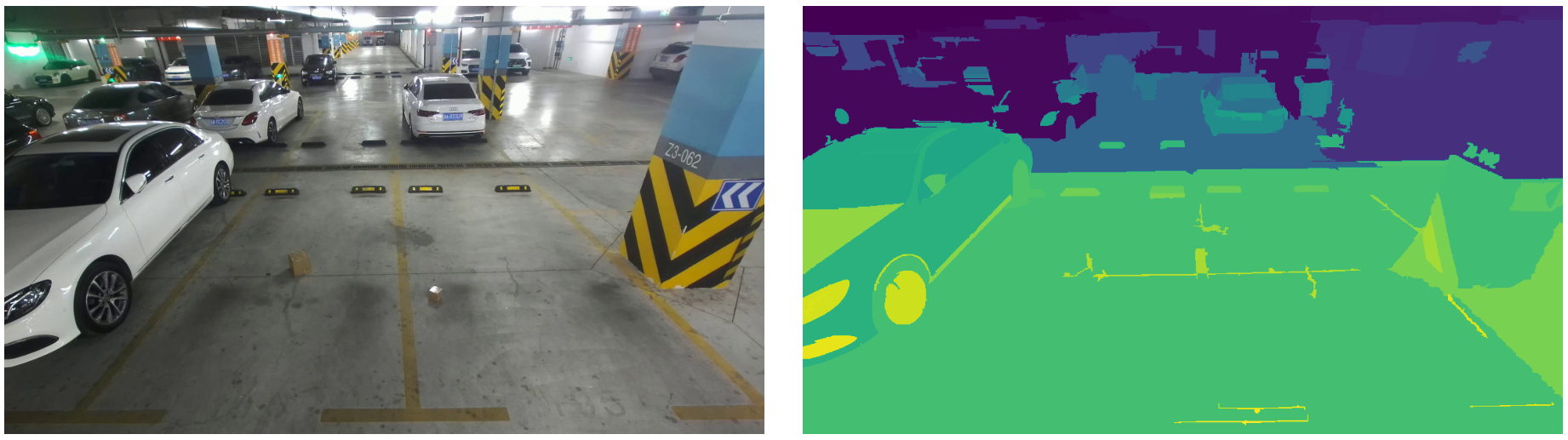}
    \caption{\fontsize{8}{12}\fontfamily{ptm}\selectfont Original test image (left) and its graph segmentation with implemented color map (right) without pre-processing}
    \label{fig:fig2}
\end{figure}

The Figure \ref{fig:fig2} shows the original test image (left) and its graph segmentation with the implemented color map (right) without preprocessing. Note that hereinafter we use the OpenCV implementation of graph segmentation with the following parameters: $sigma = 0.6$, $k = 1074$ and $min\_size = 185$. Figure \ref{fig:fig2} demonstrates that the floor is segmented quite well. However, only the larger box from two target objects (light brown boxes) that are used here as obstacles, is segmented. But even this box is faded in the image with some cracks. The small box is not segmented at all.

Among the shortcomings of the method for our task, we can note its over-sensitivity to some local changes in light and color, when some random cracks and shadows can be identified by the algorithm as separate segments. To avoid this shortcoming, we implemented image pre-processing, which consists of several steps.

At the first stage, we tried to make local changes in color and light smoother. For this purpose we blended the image with its inverse copy using the so-called soft light blending (the equation \ref{eq2}) \cite{Wiki_Blend_modes,Pegtop_softlight_mode}. To avoid the discontinuity at the 50\% gray level that is inherent in the main method, the Pegtop formula was used \cite{Pegtop_softlight_mode}.

\begin{equation}\label{eq2} 
f_{pegtop}(a,b) = (1-2b)a^{2}+2ba,
\end{equation}

where $a$ is the pixel intensity value for the image being processed, and $b$ is the pixel intensity value for its inverse copy, $b = 255 - a$. Both images are of type $uint8$ and all image pixels are in the range of intensities from 0 to 255.

\begin{figure}[!htbp]
    \centering
    \includegraphics[width=\linewidth]{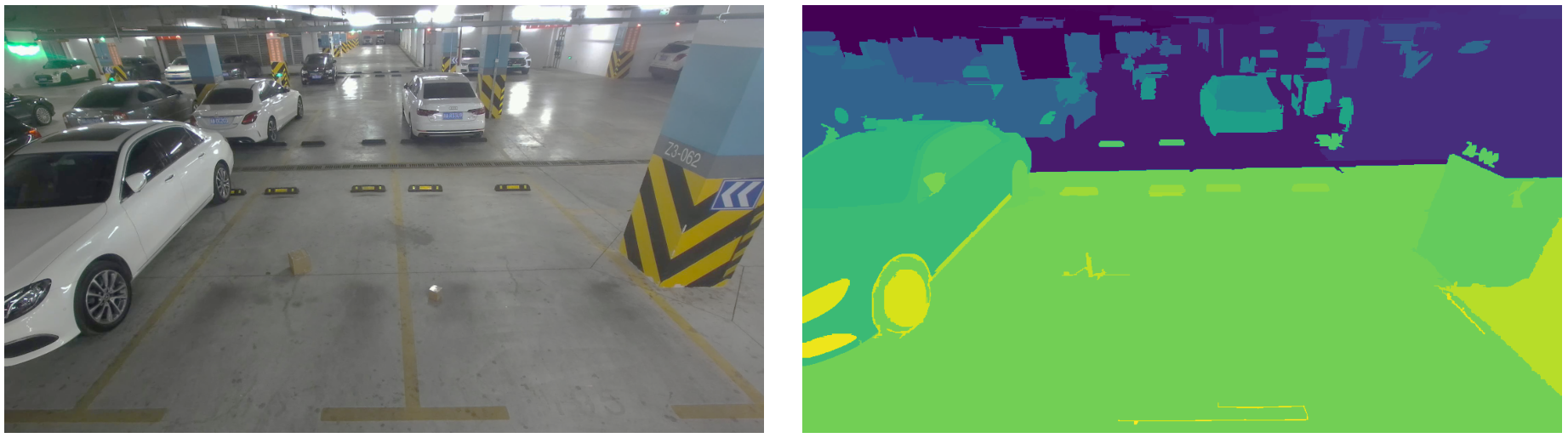}
    \caption{\fontsize{8}{12}\fontfamily{ptm}\selectfont Original test image (left) and its graph segmentation visualized as the colormap (right) using Pegtop soft light blending with an inverse copy of the image}
    \label{fig:fig3}
\end{figure}

Now that we have applied the Pegtop soft light blending, we can see in Figure \ref{fig:fig3} the changes in the segmentation. The small box is still not segmented, but the big number of cracks in the large box have disappeared, as well as a part of the shadow near the column.

\begin{figure}[!htbp]
    \centering
    \includegraphics[width=\linewidth]{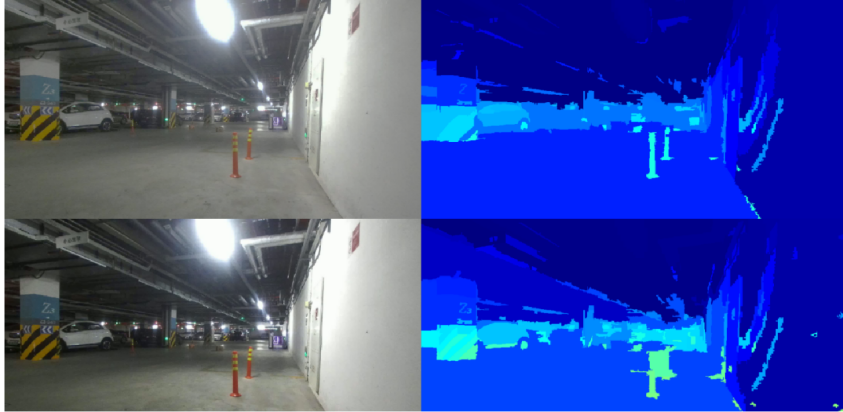}
    \caption{\fontsize{8}{12}\fontfamily{ptm}\selectfont Original images (left) and images with the results of their graph segmentation (right). The bottom images are before pre-processing, and the top image - after Pegtop soft light blending (with its inverse copy) applied to the bottom original image}
    \label{fig:fig4}
\end{figure}

The Figure \ref{fig:fig4} shows another example of soft light blending with an inverse copy of the image with the same segmentation settings. The finally segmented floor area after such pre-processing is detected at a greater distance and does not fade with the wall. In addition, road poles that are located together in the foreground of the image are correctly segmented from each other.

\begin{figure}[!htbp]
    \centering
    \includegraphics[width=\linewidth]{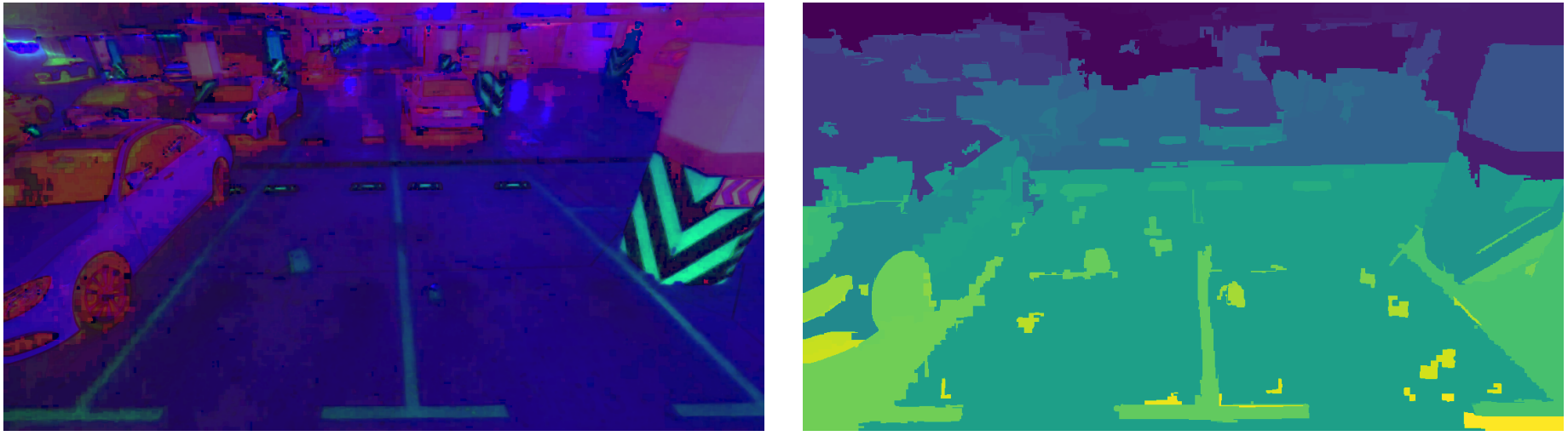}
    \caption{\fontsize{8}{12}\fontfamily{ptm}\selectfont Original test image after RGB to HSV color space conversion (left) and its graph segmentation visualized as the colormap (right)}
    \label{fig:fig5}
\end{figure}

At the second step of pre-processing, the conversion from the RGB color space to the HSV color space (the equation \ref{eq3}) was performed (see, the Figure \ref{fig:fig5}, left), since in most cases this provides better image contrast.

\begin{equation}\label{eq3} 
\footnotesize
\left \{ \begin{matrix}
V \leftarrow max(R, G, B)\\
S \leftarrow \left \{ \begin{matrix} 
V - min(R, G, B)  & if  & V \neq  0 
\\0                 &     & otherwise \end{matrix}
\right.
\\ H \leftarrow \left \{ \begin{matrix}
60(G-B)/(V - min(R, G, B))      & if  $ $ V = R 
\\120+60(B-R)/(V - min(R, G, B))  & if  $ $ V = G 
\\240+60(R-G)/(V - min(R, G, B))  & if  $ $ V = B
\\0                               & if  $ $ R = G = B
\end{matrix}
\right.
\\if H < 0                       $ $ then  $ $ H \leftarrow H+360
\end{matrix} \right.
\end{equation}

where $R$, $G$ and $B$ are red, green and blue color space components respectively; $H$, $S$ and $V$ are Hue, Saturation and Value color space components, which are in the range of\\ 0 $\leqslant H \leqslant$ 360, 0 $\leqslant S \leqslant$ 1, and 0 $\leqslant V \leqslant$ 1.

\begin{figure}[!htbp]
    \centering
    \includegraphics[width=\linewidth]{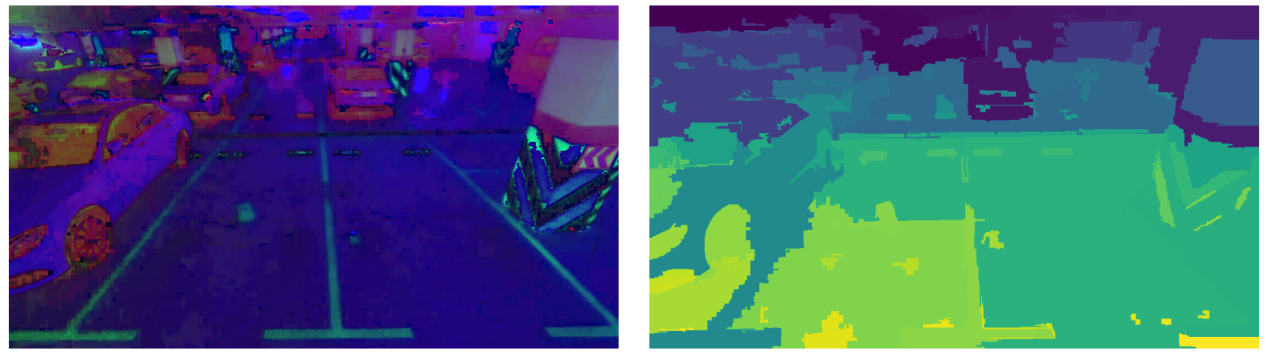}
    \caption{\fontsize{8}{12}\fontfamily{ptm}\selectfont Original test image after increasing the saturation in the HSV color space conversion (left) and its graph segmentation visualized as the colormap (right)}
    \label{fig:fig6}
\end{figure}

Both target boxes are now segmented, however there are more unwanted segments in the image, especially in the floor area (Figure \ref{fig:fig5}, right). After increasing the saturation by 50\%, the image contrast improved even more (Figure \ref{fig:fig6}, left). The quality of the segments for both boxes increased, and some unnecessary segments on the right side disappeared, but new large segments appeared on the left side (Figure \ref{fig:fig6}, right).

\begin{figure}[!htbp]
    \centering
    \includegraphics[width=\linewidth]{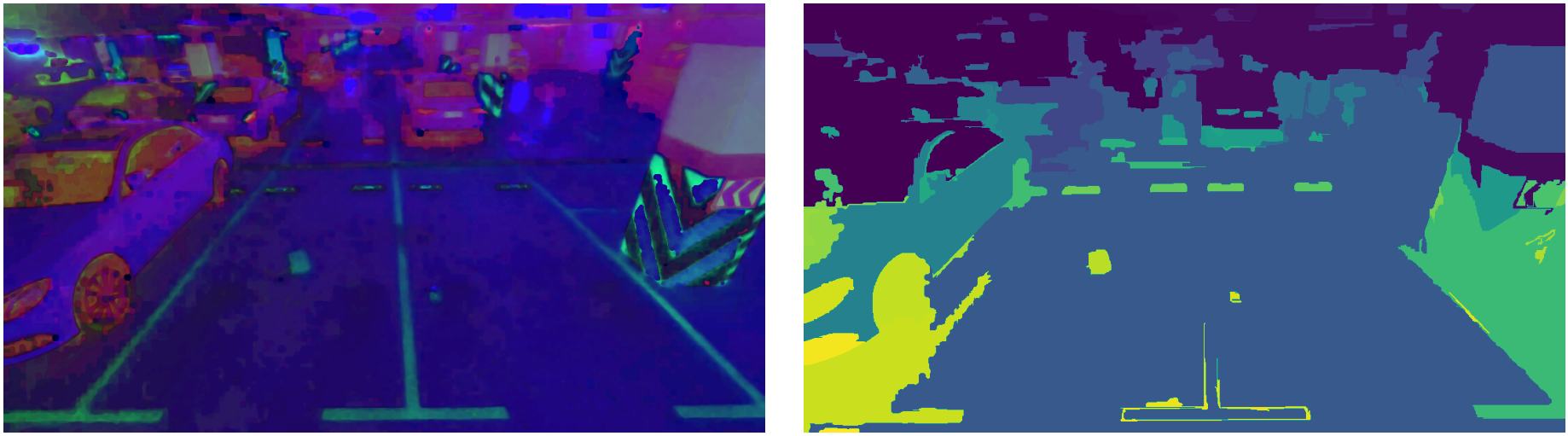}
    \caption{\fontsize{8}{12}\fontfamily{ptm}\selectfont Original test image after increasing the saturation in the HSV color space conversion and image blurring (left) and its graph segmentation visualized as the colormap (right)}
    \label{fig:fig7}
\end{figure}

 Although the graph segmentation filter has some parameters to adjust the size and number of segments, median filtering was performed to avoid false segments due to small cracks and spots on the ground. The Figure \ref{fig:fig7} shows the results of this step. The kernel size in this case is 5×5. The boxes now have fairly clean individual segments, and the number of incorrect segments is quite small.

\begin{figure}[!htbp]
    \centering
    \includegraphics[width=\linewidth]{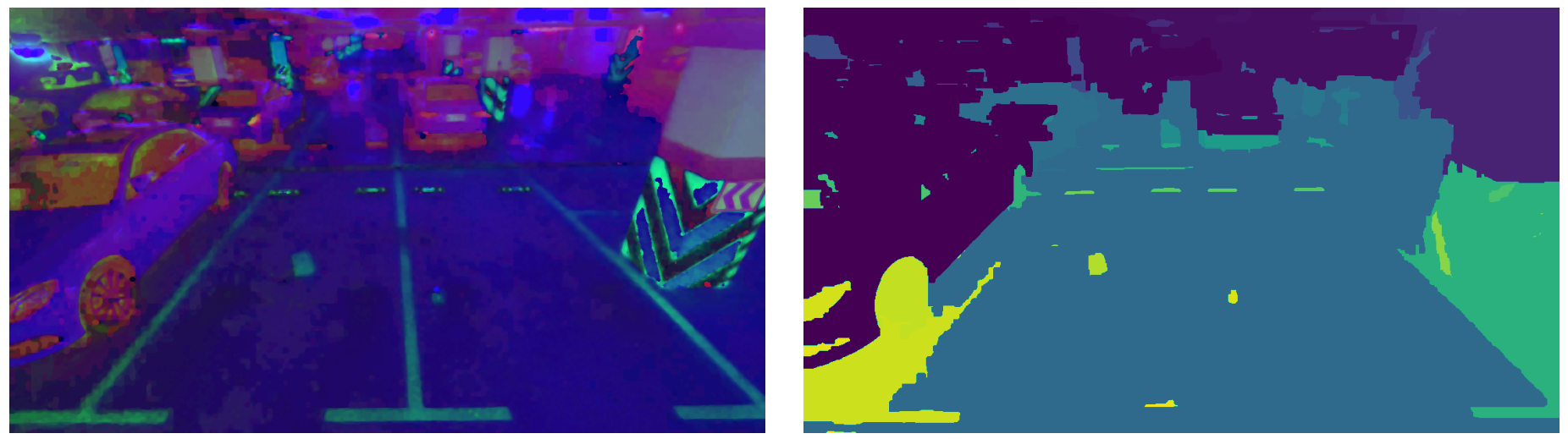}
    \caption{\fontsize{8}{12}\fontfamily{ptm}\selectfont Original test image after increasing the saturation in the HSV color space conversion and image blurring (left) and its graph segmentation after morphological erosion of the segmented image shown as a colormap (right)}
    \label{fig:fig8}
\end{figure}

However, to reduce further the number of unwanted segments, morphological erosion with a fairly large kernel (7x7 in our case) was implemented in the segmented image, as shown in Figure \ref{fig:fig8}. It is a possible optional step (that is why we did not show it on the workflow in Fig. \ref{fig:method}) that can be applied to the segmented image (\ref{fig:fig7}, right) if there is a lot of small spots.

\begin{figure}[!htbp]
    \centering
    \includegraphics[width=\linewidth]{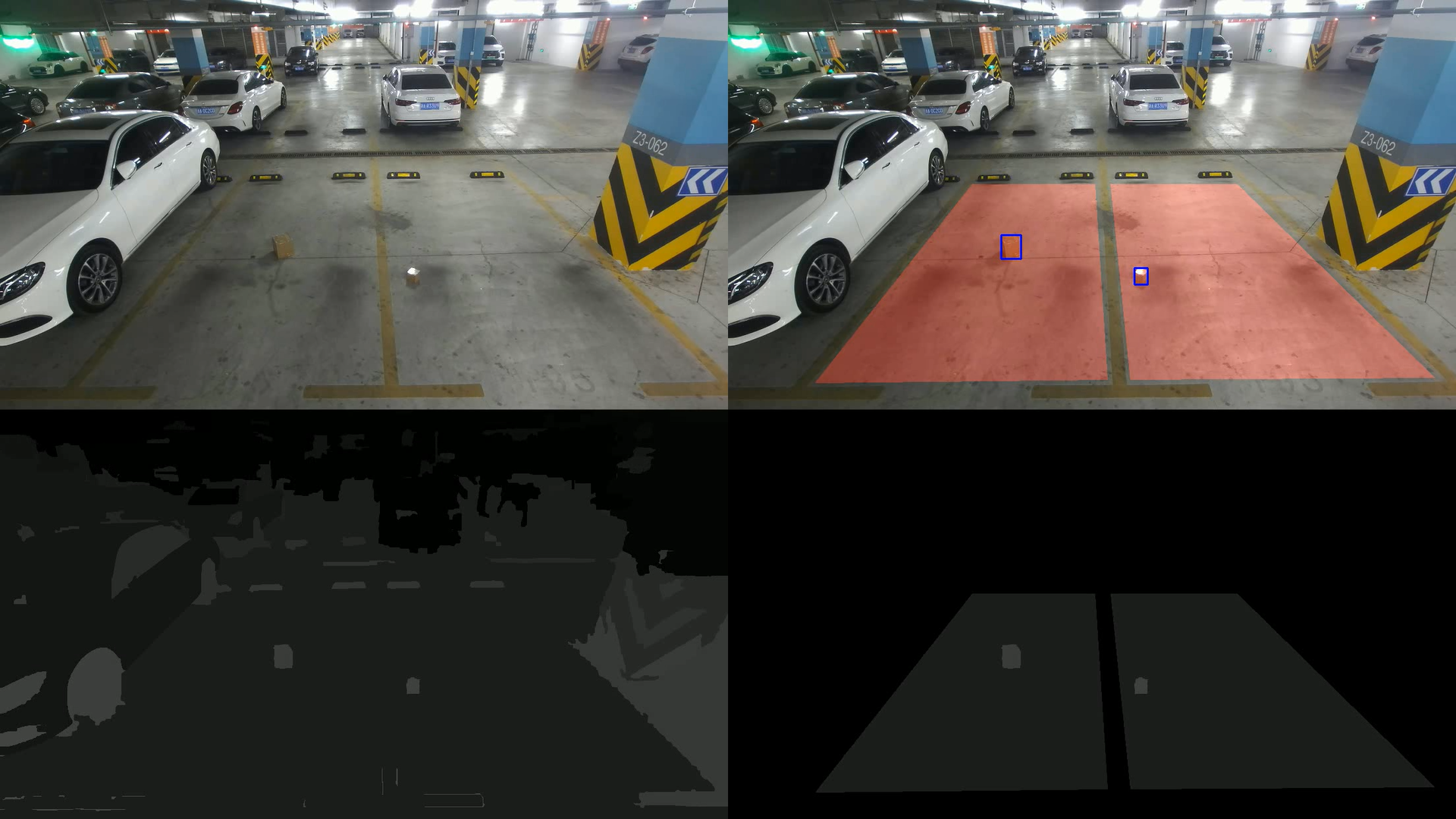}
    \caption{\fontsize{8}{12}\fontfamily{ptm}\selectfont Original test image (top left), and its segmented image (bottom left), the segmented image with selected ROI (bottom right), and the result of obstacle detection (top right)}
    \label{fig:fig9}
\end{figure}

Then the segments can be clustered using DBScan, which is a popular density-based clustering algorithm, or even found by looking for unique values. To improve the recognition rate and reduce the number of false alarms caused by cracks and marks on the ground, it is recommended to manually select the region of interest (ROI) that does not include stationary objects, road passages and parking lanes. The result of the algorithm implementation with selected ROI is shown in Fig. \ref{fig:fig9}. The red color in the upper right figure indicates that obstacles are being detected.

\subsection{Stereo-based obstacle detection}

\begin{figure}[!htbp]
    \centering
    \includegraphics[width=\linewidth]{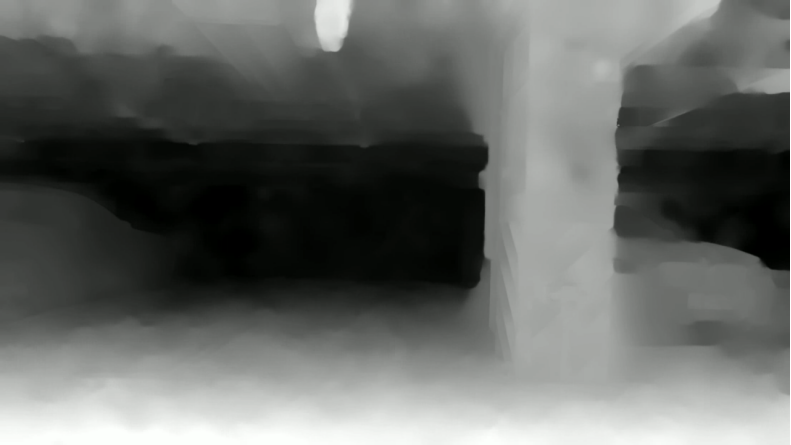}
    \caption{\fontsize{8}{12}\fontfamily{ptm}\selectfont The example of fluctuations in disparity map obtained from Stereolabs ZED 2 camera}
    \label{fig:fig10}
\end{figure}

While RGB image segmentation techniques are quite robust when they detect even very small objects, they are still limited by contrast of object to background.
For better detection of obstacles with low contrast to the background, a disparity map generated by a stereo camera was used. This method also has some limitations. First, the accuracy of the stereo image is directly proportional to the distance, the maximum value of which depends on the baseline between the cameras. In our case, the baseline is about 15 cm, so the maximum distance is limited to about 10 m, and the best accuracy is in the range of 1.5-4 m. Second, all disparity map acquisition methods are not very accurate in the presence of structureless and mirror surfaces, therefore, in this case, the disparity map appears to fluctuate on these surfaces, making smaller obstacle detection harder or even impossible (Figure \ref{fig:fig10}).

To improve the accuracy of obstacle recognition, we used the disparity map averaging by relevant RGB image segments, obtained using SLIC segmentation \cite{achanta2012slic}. To improve the accuracy of obstacle recognition, we used the disparity map averaging by relevant RGB image segments, obtained using SLIC segmentation [3]. In other words, in this particular case, the RGB image is used as a source for for disparity map segmentation. The positive feature of most superpixel segmentation methods is that they provide high-quality detection of object edges. It also speeds up the next steps because you can work with fewer superpixels instead of working with all the pixels in the image, which can be very slow at higher resolutions.


Although the disparity map contains depth information,  we intend to find 3D clusters therefore we need to convert it to 3D point cloud by converting UVZ coordinates to XYZ coordinates.
An obstacle can be defined as a volume above the ground level, so to search for such an object, we should find the ground points, and detect all points above this level that represent an obstacle in our case. To find the ground level, which is usually a plane, we use RANSAC algorithm  \cite{fischler1981} that is a robust model fitting method to detect the ground using a plane model.

\begin{figure}[!htbp]
    \centering
    \includegraphics[width=\linewidth]{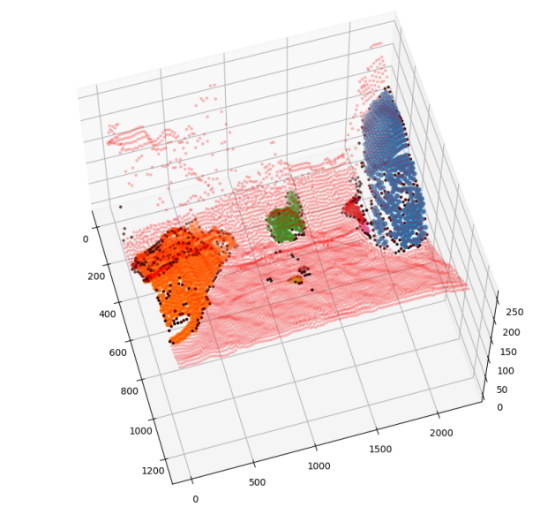}
    \caption{\fontsize{8}{12}\fontfamily{ptm}\selectfont Point cloud after RANSAC and DBScan clustering}
    \label{fig:fig11}
\end{figure}

Point cloud clustering is performed by DBScan \cite{ester1996} as this algorithm is very efficient for clustering points by density. In Figure \ref{fig:fig11}, the red 3D superpixels represent the result of RANSAC clustering, and other cluster colors show various obstacles in 3D space.

\begin{figure}[!htbp]
    \centering
    \includegraphics[width=\linewidth]{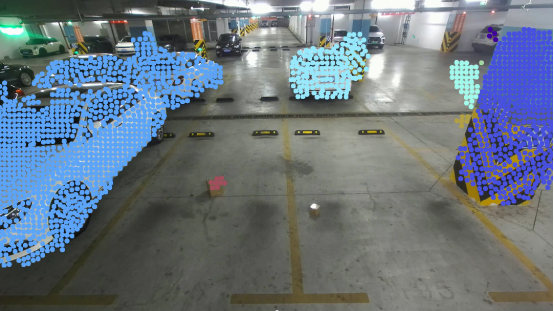}
    \caption{\fontsize{8}{12}\fontfamily{ptm}\selectfont The results of object segmentation by stereo image-based method with SLIC superpixel segmentation}
    \label{fig:fig12}
\end{figure}

In Figure \ref{fig:fig12}  the clusters of found obstacles projected onto the image. The large objects, such as cars and big boxes are detected robustly, however the smaller boxes are faded with a floor cluster due to the poor quality of the disparity map.

\subsection{Fusion}


In order to take advantage of both methods: searching small contrast obstacles in the RGB-based method and finding larger obstacles without the influence of contrast, we fused the bounding boxes from both algorithms using non-maximum suppression by central point distance. This will allow us to gain the obstacle detection rate in the scene, using both stereo and image analysis techniques.

%% file: chapters/setup.tex
\section{Experimental Setup, Tests and Results}
\label{sec:setup}

We conducted various experiments in the region of interest of the underground parking with several types of objects, such as boxes of different sizes (10x10x10cm and 20x20x20cm), orange-white traffic cones and, sometimes, a person in the camera's field of view. Thus, we collected our own dataset for such objects by moving the boxes in space and placing them on the floor at different angles to the camera, and sometimes throwing them into the camera's field of view so that they rolled along the floor at low speed.

\begin{figure}[!htbp]
        \captionsetup{font=scriptsize}
		\captionsetup[subfigure]{aboveskip=4pt,belowskip=0pt}
		\begin{subfigure}[b]{0.255\textwidth}
		    \captionsetup{font=scriptsize}
			\centering
		\center{\includegraphics[width=\textwidth]{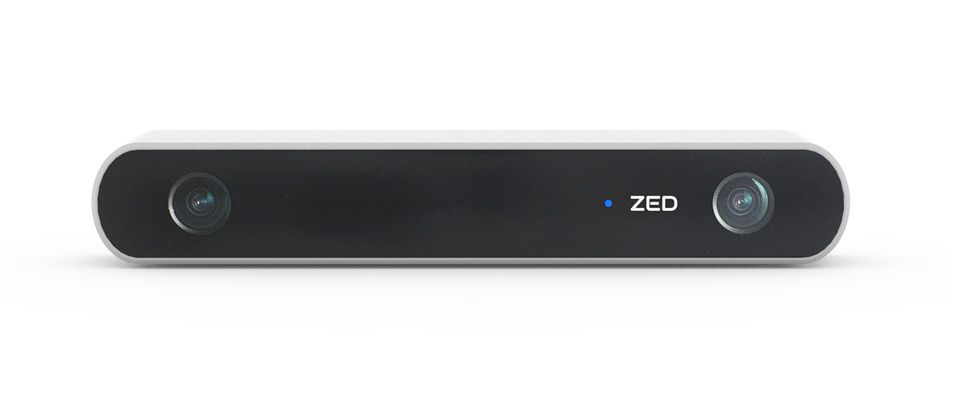}} 
			\caption{\textnormal {\fontsize{8}{12}\fontfamily{ptm}\selectfont Stereolabs ZED 2}}
			\label{fig:chassis_a}
		\end{subfigure}
		\hfill
		\begin{subfigure}[b]{0.215\textwidth}
		    \captionsetup{font=scriptsize}
			\centering
		\includegraphics[width=\textwidth]{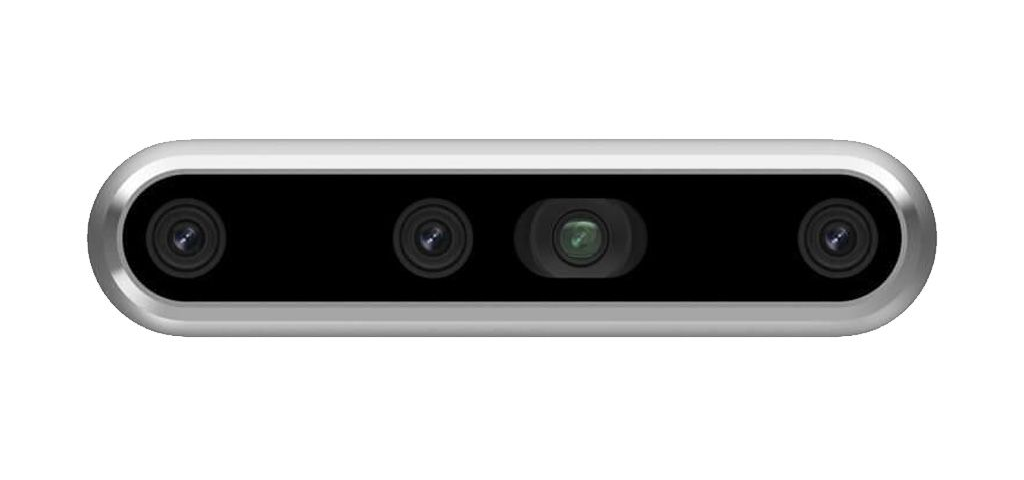} 
			\caption{\textnormal {\fontsize{8}{12}\fontfamily{ptm}\selectfont Intel RealSense D455}}
			\label{fig:chassis_b}
		\end{subfigure}
	\caption{\fontsize{8}{12}\fontfamily{ptm}\selectfont The stereocameras that we used to collect the road obstacle datasets in a parking lot scene}
	\label{fig:fig13}
\end{figure}

For our experiments on road obstacle detection, we used the Intel RealSense and Stereolabs Zed 2 stereo cameras available on the market (the stereo cameras' photo in the Figure \ref{fig:fig13} and characteristics in the Table \ref{tab02}). Intel RealSense Depth Camera D455 is the stereoscopic camera for both indoor and outdoor applications with possibility to observe the guaranteed measuring range from 0.6 to 6 m (in some cases, over 10m that varies with lighting conditions) \cite{depth-camera-d455, depth-camera-d455-datasheet}. For Intel RealSense camera the dataset consisted of left and right grayscale images, RGB images for a color camera and disparity map, obtained with camera' SDK. Intel RealSense SDK 2.0 provides an on‑chip self‑calibration option that allows D455 stereo camera calibration in less than 15 seconds without the need for specialized targets \cite{depth-camera-d455}. Stereolabs Zed 2 is also the the stereoscopic camera for indoor and outdoor applications with 120° Wide-Angle Field of View, and the measuring depth range of 0.2 to 20 m (since it has wider baseline) \cite{stereolab-ZED2, stereolab-ZED2-datasheet}. Its SDK has the lightweight neural network for stereo matching that can bring some benefits for stereo depth sensing, spatial object detection and Positional Tracking \cite{stereolab-ZED2}. For Stereolabs ZED 2 camera dataset includes video streams from left and right RGB cameras and a disparity map, retrieved from stereo camera SDK.

Since we could not adjust the baseline (the distance between the cameras), the obstacle detection range is limited to 10-15 meters.   Due to the significant stereo camera noise in the raw data streams from the cameras' SDK, the use of the stereo algorithm alone without RGB camera processing (discussed in the Section \ref{sec:methodology}) is not sufficient to distinguish objects smaller than 10x10x10cm.

\begin{table}[!htbp]
    \centering
    \caption{\textsc{Depth camera specification for Intel RealSense D455 \cite{depth-camera-d455-datasheet} and Stereolabs ZED 2 \cite{stereolab-ZED2-datasheet} \label{tab02}}}
    \begin{tabular}{|c | c|}
    \hline
    \textbf {Intel RealSense Depth Camera D455} & \textbf {Configuration} \\
    \hline
    Stereo depth resolution &     2x (1280x720) @30fps \\
    Depth streaming         &  up to 90 fps \\
    Depth Field of View (H × V)     &     87° × 58° \\
    Depth range             &   0.6 to 6 m \\
    Depth Accuracy          &   $<$2\% up to 4m \\
    Baseline                &   95 mm \\ 
    RGB streaming resolution          &     1920x1080 @30fps \\
    RGB sensor FOV (H × V)  &     90° × 65° \\
    \hline
    \hline
    \textbf{Stereolabs Depth Camera ZED 2} & \textbf{Configuration} \\
    \hline
    Stereo depth resolution &     2x (2208x1242) @15fps \\
                            &     2x (1920x1080) @30fps \\
                            &     2x (1280x720) @60fps \\
                            &     2x (672x376) @100fps \\
    Depth Field of View (H × V)    &  110° x 70° \\
    Depth Range             &   0.2 - 20 m \\ 
    Depth Accuracy          &   $<$1\% up to 3m \\
                            &   $<$5\% up to 15m \\
    Baseline                &   120 mm \\ 
    \hline
    \end{tabular}
\end{table}

The experiments were carried out on a conventional computer with an integrated graphics card, the characteristics of which are given in the Table \ref{tab01}.

\begin{table}[!htbp]
\centering
\caption{\textsc{Specifications of Machine \label{tab01}}}
\begin{tabular}{|c | c|}
\hline
  \textbf{System} & \textbf{Configuration} \\ [0.6ex] 
 \hline 
 Operating System & Windows 10 \\ 
 Processor & 3.2 GHz Intel core i7-8700\\ 
 Video adapter & Intel UHD Graphics 630 (350 - 1200 MHz) \\ 
 RAM & 16 GB \\ 
 Hardware memory & 1000 GB \\ 
 DirectX version & 12 \\
 OpenGL version & 4.5 \\
 \hline
\end{tabular}
\end{table}
 
\begin{figure}[!htbp]
    \centering
    \includegraphics[width=\linewidth]{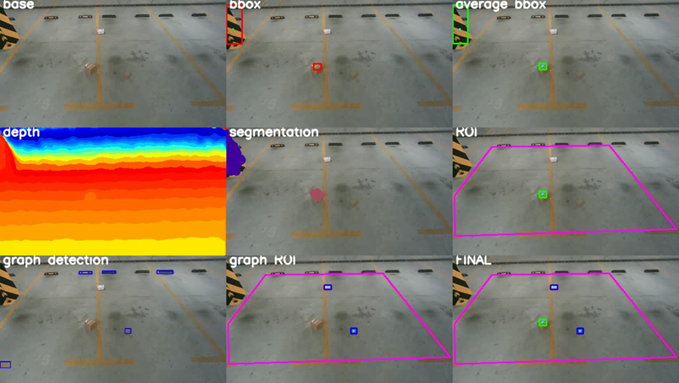}
    \caption{\fontsize{8}{12}\fontfamily{ptm}\selectfont The results of test frame processing using our dual channel obstacle detection method in a parking lot scenario with three boxes as obstacles. The bottom row shows the work of the 1st channel of the RGB image-based obstacle detection (the detected boxes are in blue color), and the right column represents the 2nd channel of stereo camera-based obstacle detection (the detected box is in green color). The bottom right picture demonstrates the result of Depth and Image Fusion where all boxes are detected with the combination of dual channel processing that increases the robustness of obstacle detection. The region of interest is shown in pink color}.
    \label{fig:fig14}
\end{figure}

\begin{figure}[!htbp]
    \centering
    \includegraphics[width=\linewidth]{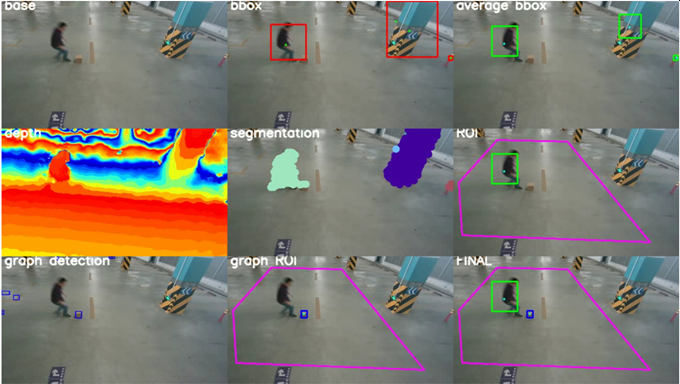}
    \caption{\fontsize{8}{12}\fontfamily{ptm}\selectfont The results of test frame processing using our dual channel obstacle detection method in a parking lot scenario with a moving box and a person as obstacles. The bottom row shows the work of the 1st channel of the RGB image-based obstacle detection (the detected box is in blue color), and the right column represents the 2nd channel of stereo camera-based obstacle detection (the detected person is in green color). The bottom right picture demonstrates the result of Depth and Image Fusion where all obstacles are detected. The region of interest is shown in pink color}
    \label{fig:fig15}
\end{figure}

 Let's analyze the results of field experiments in a parking lot scenario.
 The Figure \ref{fig:fig14} shows the intermediate and final results of frame-to-frame processing with our dual channel obstacle detection method for three boxes of different sizes as obstacles. The Figure \ref{fig:fig15} demonstrates the results of video processing for a moving box and a moving person that were taken as target objects. The green bounding boxes in the 'FINAL' frames (right bottom pictures) in Fig. \ref{fig:fig14} and \ref{fig:fig15} are the result of the stereo camera-based obstacle detection, whereas the blue ones are the RGB image-based obstacle detection. Here, you can see the interesting example when each channel detects definite obstacles, whereas combining them into Depth and Image Fusion method helps to detect each obstacle in the region of interest (ROI) that increases the robustness of obstacle detection.
 
 Let us comment in details the processing steps in the frames of Figures \ref{fig:fig14} and \ref{fig:fig15}, focusing on the text annotations in the upper left corner of each picture:
\begin{itemize}
    \item 'base' is the image from the left camera of the stereopair (Intel RealSense Depth Camera D455); 
    \item 'depth' is the depth map; 
    \item 'segmentation' is the results of the subsequent SLIC, RANSAC and DBSCAN clustering; 
    \item 'bbox' is the bounding boxes around separate clusters of the subsequent SLIC , RANSAC and DBSCAN clustering; 
    \item 'average bbox' - the results of averaging 'bboxes' over 5 frames (that was done to increase robustness of moving obctacle detection); 
    \item 'graph detection' - bounding boxes around separate segments based on  RGB image-based segmentation, 
    \item 'ROI' shows the selected region of interest (ROI) in pink color that was used as a mask to compute there bounding boxes around separate clusters of subsequent SLIC, RANSAC and DBSCAN clustering and bounding boxes averaged over 5 frames; 
    \item 'graph ROI' demonstrates the bounding boxes around separate segments of RGB image-based segmentation with selected ROI and bounding boxes averaged over 5 frames; 
    \item 'FINAL' is the fusing result of ‘ROI’ and ‘graph ROI’ bounding boxes that shows the outcome of our Depth and Image Fusion method.
\end{itemize}

Since the performance of RGB image-based obstacle detection approach was not optimized well and takes more time than stereo camera-based obstacle detection, we tested the obstacle detection with Depth and Image Fusion method for moving objects at low speeds.

%% file: chapters/conclusion.tex
\section{Conclusions and Discussion}
\label{sec:conclusion}

\subsection{Conclusions}

In this paper, we presented a methodology for dual detection of obstacles on the road recorded with Stereolabs Zed 2 or Intel RealSense D455 depth camera. We have shown with case study for detection objects with sizes of 10x10x10cm and 20x20x20cm that combining sensory information for both depth data and video, provides more accurate detection, positioning and characterization of objects. The depth and image fusion algorithm developed with the methodology demonstrates the advantages of dual detection of road obstacles using various stereo and RGB  processing channels that complement each other.

To detect obstacles on the road using RGB streaming with frame-by-frame image analysis, the OpenCV graph segmentation technique was used, followed by pre-processing: converting an RGB image to HSV with subsequent saturation, median filtering, erosion, and dilation to remove noise. For stereo-based obstacle detection, we use disparity map reconstruction from stereo (performed automatically by  selected stereo camera SDK), get a dense point cloud, and then apply a superpixel algorithm based on SLIC segmentation to speed up the point cloud processing, compute a floor plane using the RANSAC algorithm and detect objects using 3D DBSCAN clustering.

Since we have the significant background noise, which fluctuates greatly with time and illumination in the raw data streams from the SDK (we observes it for both Stereolabs Zed 2 or Intel RealSense D455 depth camera output data, and suppose that it goes from disparity map estimation), we could not distinguish between objects smaller than 10x10x10 cm without the special processing of RGB channel. Thus, the applying dual detection of obstacles on the road with the developed depth and image fusion algorithm gives encouraging results.

The main research contribution of this paper:
\begin{itemize}
    \item A stereo camera-based depth and image fusion algorithm for road obstacle detection that take advantages in searching small contrast objects by RGB-based method and finding larger obstacles without the influence of contrast  by stereo image-based approach with SLIC superpixel segmentation.
    \item An improved road obstacle detection technique using a monocular camera with frame-by-frame image analysis that applies the OpenCV graph segmentation after the preliminary processing: converting an RGB image to HSV color space with saturation increase, median filtering, morphological erosion and dilation, instead of using neural networks and a pre-recorded scene.
\end{itemize}

\subsection{Discussion}
In some cases, when the angular size of the detected object is close to the size of all sorts of different random cracks or spots, in the first part of the algorithm (by image) there may occur false positive alarms of system. Since they are not constant, but for units of frames, in general one can try to filter them by detection time, which, in its turn will for sure affect the quality and detection rate of objects, actually being in the camera frame.

Although the presented method has only been tested in a parking scenario at relatively short distances, it is reasonable to assume, based on the theory of stereo vision and epipolar geometry, that the method can be used at longer distances by increasing the baseline for stereo camera (which is only about 15 cm in our case) and focal length.
In our further research we suppose to compare this detection method with existing ones in terms of accuracy, algorithm performance, and robustness to the various object detection.